\def\year{2019}\relax
\documentclass[letterpaper]{article} 
\usepackage{aaai19}  
\usepackage{times}  
\usepackage{helvet} 
\usepackage{courier}  
\usepackage[hyphens]{url}  
\usepackage{graphicx} 
\def\UrlFont{\rm}  
\usepackage{graphicx}  
\usepackage{url}

\usepackage{color}
\usepackage{soul}
\usepackage{enumitem}

\newcommand{\code}[1]{{\texttt{#1}}}

\title{Metrology for AI: From Benchmarks to Instruments}
\author{Chris Welty, Praveen Paritosh, Lora Aroyo\\Google Research} 

\begin{document}

\maketitle

\begin{abstract}
In this paper we present the first steps towards hardening the science of measuring AI systems, by adopting metrology, the science of measurement and its application, and applying it to human (crowd) powered evaluations.  We begin with the intuitive observation that evaluating the performance of an AI system is a form of measurement. In all other science and engineering disciplines, the devices used to measure are called instruments, and all measurements are recorded with respect to the characteristics of the instruments used.  One does not report mass, speed, or length, for example, of a studied object without disclosing the precision (measurement variance) and resolution (smallest detectable change) of the instrument used. It is extremely common in the AI literature to compare the performance of two systems by using a crowd-sourced dataset as an instrument, but failing to report if the performance difference lies within the capability of that instrument to measure. To illustrate the adoption of metrology to benchmark datasets we use the word similarity benchmark WS353 and several previously published experiments that use it for evaluation.
\end{abstract}

\section{Contributions of this paper}
In this paper we examine the question of how the variations in human interpretation and other aspects of data collection can affect the measurements we make with crowd-powered datasets. For this, we adopt metrology, the science of measurement and its application, and apply it to human (crowd) powered evaluations. 

We begin with the intuitive observation that evaluating the performance of an AI system is a form of measurement. In all other science and engineering disciplines, the devices used to measure are called instruments, and all measurements are recorded with respect to the characteristics of the instruments used.  One does not report mass, speed, or length, for example, of a studied object without disclosing the precision (measurement variance) and sensitivity (smallest detectable change) of the instrument used. It is extremely common in the AI literature to compare the performance of two systems by using a crowd-sourced dataset as an instrument, but failing to report if the performance difference lies within the capability of that instrument to measure. 

In this paper, to illustrate the adaptationn of metrological characteristics to crowd-powered instruments, we use as an example WordSim (WS353) \cite{finkelstein2002placing}, a widely used benchmark for measuring a system's ability to compute similarity between two words. This dataset has been cited over 1500 times, and has spurred the development and evaluation of automated approaches to computing lexical/semantic similarity \cite{witten2008effective,agirre2009study} and word embeddings \cite{mitchell2008vector,mikolov2013efficient,levy2015improving,pennington2014glove,bojanowski2017enriching}. 

Our contributions are:
\begin{itemize}
    \item adopting metrology for characterizing crowdsourced datasets 
    \item a guide for turning benchmark datasets into instruments
    \item hardening the science of measuring AI system
    \item increasing awarness on the impact of benchmark characteristics on the reproducibility and reliability of AI results
\end{itemize}

\section{Benchmarks are Instruments}
To evaluate AI systems that interpret unstructured input (e.g. image, audio, video, text, etc.) there are commonly used benchmark datasets that have been regarded as standards for evaluation. In AI, this is by and large the only method of evaluating AI systems -- by comparison to humans performing "the same" task. Due to the fact that it takes significant effort and resources to create an annotated corpus, existing datasets become benchmarks through \textit{timeliness} (i.e. being first to provide open access to such dataset), or \textit{size} (i.e. being the largest dataset for a specific task). Once a dataset has been 'established' as a benchmark due to its frequent use, researchers typically do not question its \textit{quality} and \textit{reliability} as an instrument. 

\cite{snow2008} introduced the power of using the crowd to collect data which previously required linguists, by showing that the crowd was producing similar results to the linguists' ground truth (“gold standard”). By presenting this sort of expert-data as ground truth, most of the early researchers and adopters of the crowd made a dangerous assumption: that the previous instrument was perfect, which the linguists never really claimed. In addition, humans have various cognitive biases that influence the way they interpret statements, make decisions and remember information. Ambiguity creates additional uncertainty in practically every facet of crowdsourcing and expert annotation. Ambiguity can result from missing details, contradictions and subjectivity, which may stem from differences in cultural context or individual perception of hard-to-quantify properties. All of these can leave annotators with varying interpretations, leading to results that some may regard as ``wrong''.

A growing number of researchers have started raising awareness on how properties of the data in benchmark datasets used for evaluation of various AI systems play a critical role in the stability of the results. Many of them have pointed out numerous problems and inconsistencies in the ways:
\begin{itemize}
    \item data in AI benchmarks is collected\cite{VanSon2018,Inel2017,Inel2019}
    \item results from the evaluations of AI systems against these benchmarks are presented\cite{Rogers2017,Gladkova2016,Wendlandt2018,Ramakrishnan2018}
\end{itemize}

According to \cite{Baker2016} we experience a ``reproducibility crisis'' across various scientific fields periodically. \cite{Crane2018} focuses on reproducibility of results in deep learning. \cite{Arguello2016} and \cite{Lin2016} have identified issues with reproducibility in the information retrieval community. The need for replicable and reproducible results has also been included in a list of challenges for the ACL (as reported by Joakim Nivre in 2017). However, none of these efforts focuses on the role of benchmark datasets and their quality as instruments for measuring system performance, and thus also impacting the reliability of results. For example, \cite{Ristoski2016} offer an overview of 22 benchmark datasets for machine learning on Semantic Web, however none of these have been evaluated based on their reliability as an evaluation instrument. Similarly, ACL Wiki maintains a state-of-the-art repository of standard evaluation methods and datasets for various core natural language processing tasks to ``encourage more effort into reproducibility of results''\footnote{\url{https://aclweb.org/aclwiki/State_of_the_art}}. Again here, there is no focus on assessing the quality of these benchmark datasets as instruments to reliably measure the quality of system performance. 

In this paper, we show that paying attention to the instrument characteristics of a benchmark affects how we make interpretations about the systems being measured, a key question of interest to AI researchers using these benchmarks. Characterizing crowd-powered datasets will tell us how to build better instruments and justify allocating resources to improving them. The job of task design is often relegated to ad-hoc templates and guidelines, which reduces the reusability of the data collected. In addition to characterizing all the widely used instruments, there are many open questions around how to better characterize such instruments. We also need to look toward the empirical and theoretical body of work from behavioural sciences regarding the principles of measurements involving humans \cite{Berglund2013}.

\section{Metrology: The Science of Measurement}

Measurement has been a key factor for progress in science. The need for reliable measurement in trade, agriculture, and construction has been recognized by mankind from ancient times. The international coordination of measurement began in the $18^{th}$ century, with the institution of decimal metric systems and platinum standards for meter and kilogram maintained in the Archives de la Republique in Paris (1799). This process continued in the $19^{th}$ century and led to the Metre Convention, a treaty that was signed in Paris in 1875 by representatives of seventeen nations. This convention also created the \textit{Bureau International des Poids et Measures} \cite{BIPM2004}, the international reference body for \textit{metrology}\footnote{\footnote{https://www.bipm.org/}}, a single coherent system of measurements throughout the world. 
Metrology helped make substantial progress in science and engineering over the past two centuries by increasing the \textit{reproducibility}, coordination, and \textit{reusability} of methodologies, tools, and data.  One does not use a meter-stick to measure the width of a piece of paper - it lacks the \textit{resolution} to tell if one page is thicker than another.  A micrometer is \textit{sensitive} to differences of .01mm, but a cheap one will have low \textit{precision} (high variance in the same measurement).  When these characteristics are reported with a measurement, we have a deeper understanding of the result.

In the middle of the $20^{th}$ century physicists and psychologists worked in parallel on the development of measurement, as they strongly disagreed on the meaning of measurement and the possibility of “measuring” sensory events. 
In psychology there are two schools of measurement. \textit{Psychometrics} aimed at the measurement of individual differences, i.e. the objective measurement of attributes of persons such as skills, knowledge, abilities, attitudes, personality traits, and educational achievement \cite{Cattell1890,Galton1887}. \textit{Psychophysics} \cite{Gescheider1997} aimed at measuring attributes of stimuli, which led to the development of experimental psychology, and standardized testing\cite{Kaplan2010}. However, progress required an attempt at convergence, i.e. interlinking of related developments across a variety of disciplines, embracing the physical, biological, psychological, and social sciences \cite{Berglund2013}. Especially, in the context of measurement with persons, which involves human perception and interpretation for measuring complex holistic quantities and qualities. Providing the means for reproducible measurement of parameters such as pleasure and pain has important implications in evaluating all kinds of products, services, and conditions. 

Fueled by emerging measurement needs in science and society, such as quality of products and services, there is increasing recognition and involvement of humans in measurement. This has led to  interdisciplinary collaboration that ensures common development. A new paradigm has recently emerged, \textit{soft metrology}, defined as \textit{measurement techniques and models which enable the objective quantification of properties which are determined by human perception,} where \textit{the human response may be in any of the five senses: sight, sound, smell, taste, and touch} \cite{Pointer2003}. 

Further, the project \textit{Measuring the Impossible} (MtI) called for measuring economic aspects, such as \textit{products and services that appeal to consumers according to parameters of quality, beauty, comfort, etc., which are mediated by human perceptions,} and  social aspects such as, \textit{public bodies such as hospitals, provide citizens with support and services whose performance is measured according to parameters of life quality, security or wellbeing} \cite{Pendrill2010}. This paves the way for our proposal of building instruments that assume the crowd as a first-class component. 

\section{Types of Crowd-Powered Instruments}
In recent years, crowds have been increasingly employed in solving variety of data annotation tasks. Erroneous crowd judgments and spam have been a major challenge in this process, and thus quality control is one of the most researched questions in crowdsourcing \cite{Lease2011}. To ensure quality of the resulted crowdsourced data, various mechanisms have been developed to measure the quality of the raters; \cite{Joglekar2015} provide concise confidence intervals on raters error rates, \cite{Eickhoff2018} focus on cognitive biases in crowdsourcing, \cite{Ipeirotis2011} offers a repeated acquisition of ``labels'' to resolve imperfect labeling. Despite the fact that numerous problems and inconsistencies have been pointed out, both in the way the quality of the crowd is assessed \cite{Joglekar2015} and in the overall process of data collection, there is still little attention paid to assessing the reliability of such annotated corpora as instruments for measuring system performance.

There are many different ways of building instruments from the crowd. Some of the more common types are:

\begin{itemize}
    \item \textbf{A/B testing:} The crowd can be used as an instrument through A/B testing, where user facing systems split users into two groups and provide input in two versions of the system. The objects being measured are the systems versions; differences in crowd behavior are recorded and used to show improvments.  These tests today rely purely on scale for statistical power and significance, however little else of these instruments are evaluated.
    \item \textbf{Polls, etc.:} The oldest and most deeply studied kind of crowd instruments are polls, surveys, questionnaires, and tests, which serve as instruments to measure people, their knowledge and attitudes.  Aggregated results of these studies can then be used as instruments to measure related issues, e.g. effectiveness of teachers, popularity of public figures, etc. Services such as The College Board (who run the US SAT) and Gallup (the international polling organization) are extremely rigorous about these instruments and their evolution is an active science. In AI systems, tests can be used as instruments to measure systems as they do people, e.g. reading comprehension exams can be used to measure AI reading systems, and surveys can provide very indirect measures of vague characteristics such as ``quality'' and ``usefulness'' of products. 
    \item \textbf{Direct evaluation:} In many cases, notably large companies and DARPA projects, the crowd is directly given system input and outputs and tasked to grade them, e.g. on a Likert scale.
    \item \textbf{Knowledge elicitation:} Often the crowd is used to gather or organize knowledge, such as WordNet, Wikipedia, Freebase, Wikidata, DbPedia, etc. In this case, there is no object being measured, but the resulting artifacts can be used as instruments to measure the performance of an AI system.  For example, in relation extraction systems (e.g., \cite{mintz2009distant}) are measured on their ability to reconstruct or extend Freebase\cite{bollacker2008freebase,kochhar2010anatomy}.  Little is known or published about e.g. Freebase as an instrument, and in general using such artifacts as an instrument can suffer from issues of granularity, ambiguity, perspective, and bias.
    \item \textbf{Annotation:} The most common basis of a crowd-powered instrument in AI is when the crowd are given inputs and tasked to provide output labels, or annotations, such as depicted objects in images, answers to questions, types of named entities, etc. In some cases these human annotations are themselves measurements of something: in the WordSim (WS353) dataset \cite{finkelstein2002placing} for example, which we discuss in detail in this paper, the annotations are measurements of similarity between a pair of words. In other cases the accumulated annotations become similar to knowledge elicitation artifacts, differing perhaps only in the intent for which the data was acquired (Freebase was not created to be a standard for measuring AI systems, and WS353 was). The annotations then form the basis for instruments that measure AI systems performing the same task, and will also have issues of granularity, ambiguity, perspective, and bias in the human collected data.  These issues have only recently become the subject of serious study, and we are aware of no previous work to characterize how accurate the annotations are as instruments. 
\end{itemize}

\section{From Benchmarks to Instruments}
Metrology can help the science of AI measurement by providing a basis of communication to enhance reproducibility, coordination, and reusability of our methodologies, tools, and data. It can also help to understand the reliability of measurements well enough for comparison, which is in effect what AI science needs from human computation.  

In this section, we take the example of WordSim (WS353)\footnote{\url{https://aclweb.org/aclwiki/WS353ilarity-353_Test_Collection_(State_of_the_art)}}, a widely used benchmark for measuring a system's ability to compute similarity between two words. This dataset has been cited over 1500 times, and has spurred the development and evaluation of automated approaches to computing lexical/semantic similarity \cite{witten2008effective,agirre2009study} and word embeddings \cite{mitchell2008vector,mikolov2013efficient,levy2015improving,pennington2014glove,bojanowski2017enriching}. 


Other datasets for word similarity exist, we focus on this dataset as our first characterization of crowd-powered instruments due to its familiarity, widespread use as an evaluation baseline, and simplicity.  WS353 also has the advantage that all the crowd data has been made available by its creators, including the task description and individual votes. This has made it possible for us to experiment with the effects of different metrological characterizations.

We borrow the framework from metrology [BIPM and BIPM/ GUM]\footnote{https://www.itl.nist.gov/div898/handbook/glossary.htm} and modify it to suit crowd-powered instruments, and list the key properties of instruments before placing trust in them.  For each property, we desribe how to apply it to the WS353 dataset.

\subsection{Measurement procedure} 
\textit{A detailed description of a measurement according to one or more measurement principles. It is usually documented in sufficient detail to enable an operator to perform a measurement. Changes to the measurement procedures, such as changing the guidelines, create a new instrument.} A common measurement procedure for crowd-powered instruments includes the definition of guidelines, examples and training material together with a user interface for a crowd task.  We suggest that all benchmark datasets should provide this information, as it facilitates replication and makes it clear what the intended use of the instrument is.  We provide below a candidate description for WS353:

\begin{itemize}
    \item \textbf{WS353 Crowd Measurement Procedure}
    \begin{itemize}
        \item a set of 353 word pairs selected using a WordNet\footnote{\url{https://wordnet.princeton.edu/}} heuristic was presented to a crowd to rate each pair on a similarity scale \code{[0,10]};  \item 13 raters rated the similarity of all 353 word pairs
        \item 16 raters rated the similarity of only 200 word pairs
        \item to calibrate and test the raters two pairs were used:
        \begin{itemize}
            \item a repeated pair \code{(money, cash)}
            \item a repeated word \code{(tiger, tiger)}
        \end{itemize}
    \end{itemize}
    \item \textbf{AI System Measurement Procedure with WS353}
        \begin{itemize}
            \item The collected data from the raters formed a reference standard (i.,e. benchmark) of word similarity, analagous in the metrological sense to the original meter rod. Measuring instruments are then calibrated to this standard and used to measure AI systems that learn to predict word similarity. 
            \item one explicit pair was typically dropped in the AI system evaluation
            \item no instructions are given with the dataset on what metric to use for AI system evaluation
            \item the most common practice for AI system evaluation was to use Spearman's rho (rank correlation) between an AI system predictions and the mean of the raters' values. The rank correlation was chosen, presumably, because the distribution of the WS353 rater means is exponentially distributed, and often on a completely different scale than machine prediction scores. 
        \end{itemize}
\end{itemize}

\subsection{Principle of measurement} 
\textit{The phenomenon serving as the scientific basis of a measurement. For example, the thermoelectric effect applied to the measurement of temperature. Changes to the measurement principles would create a new instrument.} 

Crowd-powered instruments are usually missing a scientific principle of measurement, and rely on the intuition that intelligent tasks can be done by people. Only recently we have begun to consider the role of individuals, such as bias and perspective, let alone item ambiguity and context.   We suggest that all benchmark datasets should also provide this information, as it facilitates replication and makes it clear what the intended use of the instrument is. We present an example of the principle of measurement for WS353:  

According to the creators of WS353, the word pairs in WS353 were selected from WordNet by beginning with a 'random' (human selected) base word, and then navigating up the hypernym graph \code{n} steps (where n varied), and down the hyponym graph n steps. Some pairs were then also chosen completely at random from WordNet. The intution behind this sampling process was that words chosen at random are likely to be dissimilar, and that this \code{2n} distance between some selected words should loosely correlate with similarity. 

Since the task was very subjective, the authors of WS353 expected that asking a relatively large number of raters (for its time, \textit{13-16 raters per item} was quite high) to rate each item would smooth out the subjectivity and result in a more objective reference standard. Asking people to \textit{rate word similarity on a scale of \code{0-10}} was also based on an intuition that there are many degrees of similarity, and that by giving raters more choices, they would be able to express very fine grained similarity, e.g. teasing apart the similarity of \code{(money - laundering)} from \code{(money - dollar)}. However, the authors of WS353 have not considered that different forms of priming might affect the human performance, such as giving some raters \code{(professor - botanist)} and \code{(professor - cucumber)} in succession, while other raters get \code{(money - laundering)} and \code{(professor - cucumber)}. According to the metrology definition of principle of measurement, different forms of priming changes the instrument.

\subsection{Indication} 
\textit{The quantity value provided by a measuring instrument. It can be presented in visual or acoustic form or transferred to another device. An indication is often given by the position of a pointer on the display for analog outputs, a displayed or printed number for digital outputs, a code pattern for code outputs, or an assigned quantity value for material measures.} 

In crowd-powered instruments, the result of the crowd data collection is typically measured as the mean per item, the mean plus standard deviation, the set of the individual rater scores, or the majority vote. In collecting WS353 data from the crowd to measure word similarity, the indication is the mean reported similarity from 13-16 crowd raters.

When the crowd data is used for measuring an AI system's performance, rank-correlation score, value-correlation score, precision, recall or F-measure could be used as indication. In most usage of WS353 as an instrument to measure AI system's performance, the indication is the rank-correlation score between the human mean and the system's prediction.

\subsection{True value} 
\textit{The value consistent with the definition of a given particular quantity, in other words, the value that would be obtained by a perfect measurement. The key insight is that metrology recognizes that the true value is \textbf{unknowable}.} 
\begin{itemize}
    \item In the \textit{Error Approach} to describing measurement, a true quantity value is considered unique and, in practice, unknowable.
    \item The \textit{Uncertainty Approach} recognizes that, owing to the inherently incomplete amount of detail in the definition of a quantity, there is not a single true quantity value but rather a set of true quantity values consistent with the definition. However, this set of values is, in principle and in practice, unknowable.
    \item Other approaches dispense altogether with the concept of true quantity value and rely on the concept of metrological compatibility of measurement results for assessing their validity.
    \item There are many measuring frameworks in science for which true values are unavailable, and one important approach is to provide different instruments, with different characteristics, which can be calibrated with respect to each other.
    \item Many \textit{crowd-powered instruments} for measuring AI systems will not have true values, indeed the instrument itself is widely (and inaccurately) viewed as providing it. 
    \item The most prevalent way to approach a true value for crowd instruments is to \textit{employ experts}. Experts can provide extremely useful information, particularly in specialist areas like medicine, astrophysics, law, etc., but several recent studies have uncovered the rather obvious point that, as humans, experts are fallible and make mistakes \cite{Inel2017,VanSon2018,Li2015,Dumitrache:2018b}. Even expert-crowd powered instruments must be analyzed and characterized \cite{Inel2019}.
\end{itemize}

Characterizing the true values of WS353 is impossible, however it is worthwhile to provide with any instrument a discussion of the approach to truth and the underlying risk.

For WS353, there is clearly no true value of word similarity except for the control case of a repeated word \code{(tiger - tiger)}. It is possible that no two words are completely unrelated, especially when context is considered. The lowest scoring pair in WS353 is \code{(cabbage - king)}, which to fans of Lewis Carroll or O. Henry are quite strongly related.

In the absence of ground truth, we propose measuring \textit{validity} - the extent to which the scores from a measure represent the variable they are intended to. Validity is not mentioned or defined by metrology, but is more developed in psychometrics.

\subsection{Accuracy of measurement} 
\textit{The closeness of the agreement between the result of a measurement and a true value of the measurand. Given that the true value is considered unknowable, both in practice and in principle, the concept measurement accuracy is not a quantity and is not given a numerical quantity value.} 

In crowd-powered instruments, the true value is usually unknowable. It is quite popular in the crowdsourcing and AI research to assume that some aggregation, such as mean or majority ratings from a number of crowd raters, can be used as the true value, thus allowing us to compute accuracy. However, \textit{this is not metrologically valid}, and other metrological properties, such as repeatability and reproducibility become more important, which we discuss below. 

\subsection{Precision of measurement} 
\textit{The closeness of agreement between indications or measured quantity values obtained by replicate measurements on the same or similar objects under specified conditions. Two conditions of precision, termed repeatability and reproducibility conditions, have been found necessary and, for many practical cases, useful for describing the variability of a measurement method.} According to ISO 5725 \footnote{\url{https://www.iso.org/obp/ui/#iso:std:iso:5725:-1:ed-1:v1:en}}, many different factors (apart from variations between supposedly identical specimens) may contribute to the variability of results from a measurement method, including: 
\begin{enumerate}[label=(\alph*)]
\item the operator
\item the equipment used
\item the calibration of the equipment
\item the environment (temperature, humidity, air pollution, etc.)
\item the time elapsed between measurements
\end{enumerate}

Under repeatability conditions, factors a) to e) listed above are considered constant and do not contribute to the variability, while under reproducibility conditions they vary and do contribute to the variability of the test results. Thus, \textit{repeatability and reproducibility are the two extremes of precision, the first describing the minimum and the second the maximum variability in results.} Other intermediate conditions between these two extreme conditions of precision are also conceivable, when one or more of factors a) to e) are allowed to vary, and are used in certain specified circumstances. Precision is normally expressed in terms of standard deviations.

Measurement precision is usually expressed numerically by measures of imprecision, such as standard deviation, variance, or coefficient of variation under the specified conditions of measurement. Sometimes “measurement precision” is erroneously used to mean measurement accuracy. In the next two sections, we show how to compute precision under repeatability and reproducibility conditions for WS353. 


\subsection{Repeatability (of results of measurements)} 
\textit{The component of measurement precision that is the variability in the short term, and occurs under highly controlled repeatability conditions (e.g. same metrology instrument, operator, setup, environment) This is the closeness of the agreement between the results of successive measurements of the same measurand carried out under the same conditions of measurement. Repeatability may be expressed quantitatively in terms of the dispersion characteristics of the results.} 

Crowd-powered instruments typically include a sample from population of crowd raters for repeatability measured by the inter-rater reliability of the human responses. For example, the published WS353 data contains 13 repetitions per word-pair, obtained by presenting the same pair to different raters. While utilizing repetition is a key part of crowdsourcing, the variability across repetitions is often unreported \cite{paritosh2012human}. 

Linguists, borrowing from sociologists, recognized the inherent variability in semantic tasks understood that there was no ground truth or gold standard. What mattered to them was independent agreement between multiple annotators, and iteratively revising the guidelines to produce consistent observations. Thus, a slew of inter-annotator agreement (also called inter-rater reliability, or IRR) metrics such as Fleiss' Kappa, or Cohen's pi, which was then generalized to all different scales by Krippendorff's alpha. 

We utilize Krippendorff's alpha as a summary statistic for repeatability. For the original WS353 instrument, $\alpha=0.59$. This is moderate to high agreement, typically, it is a standard practice to use a threshold of \code{0.6} for publishable data in linguistics.  

Often, improvements to IRR scores such as $\alpha$ can be achieved by updating the procedures or principles of measurement, especially guidelines.  However, over-tuning guidelines to achieve better IRR can serve to hide causes of disagreement that are better left exposed. These kinds of experiments have been published before, but for our purposes, IRR is \emph{one part} of the overall characterization presented here.

\begin{figure}[hbt!]
\centering
\includegraphics[width=1.0\columnwidth]{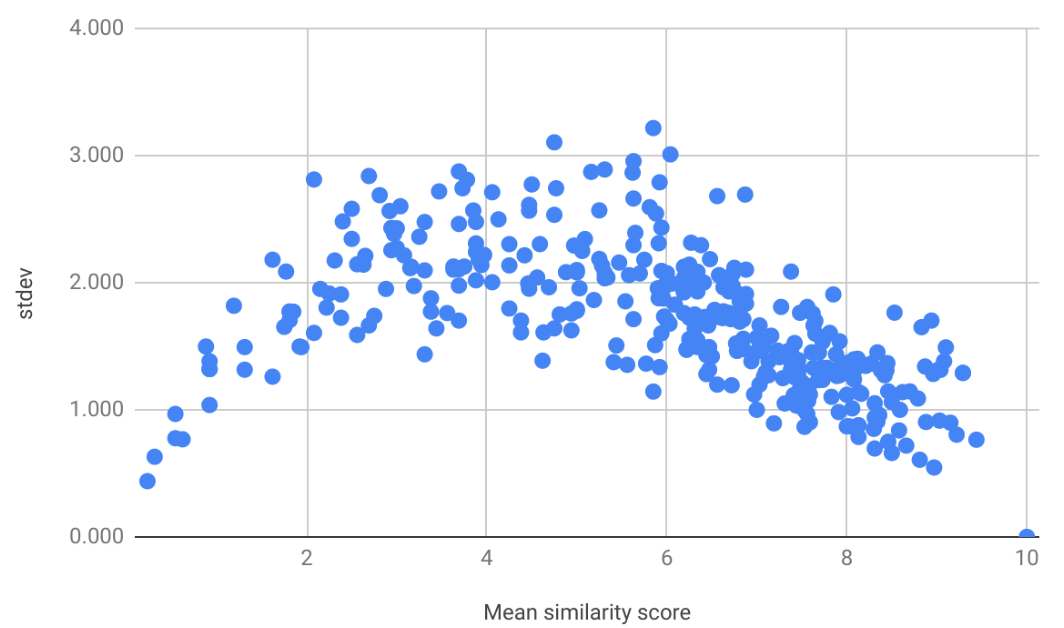}
\caption{The metrological precision of WS353 can be vizualized by comparing the standard deviation of each pair's votes to the mean.}
\label{fig:precision_per_worker_353}
\end{figure}

These repeated measurements give us a sound basis to measure the precision of the WS353 instrument, however the variance per item itself varies dramatically. Ultimately, the precision of the WS353 instrument is a vector of per-item standard deviation scores (stdev is more useful than variance as it is relative to the mean).  Overall, all the WS353 word pair standard deviations are normally distributed, with a mean standard deviation of \code{1.7}, and a standard deviation of deviation of \code{.54}. The highest variance is in the pair \code{(precedent - example)}, and the lowest aside from \code{(tiger - tiger)}, is \code{(king - cabbage)}, which also has the lowest similarity score. In general, the standard deviations have a crescent shaped relationship with the mean (Figure \ref{fig:precision_per_worker_353}), and while its mathematically necessary that low and high mean scores have low variance, there is nothing mathematically preventing a pair from having universal agreement on a mid-range score. This implies that \code{67\%} of the pairs have rater scores that vary from the mean by between \code{1.16} and \code{2.24}.  \textit{This is a low precision instrument.} 

By contrast, the standard and accepted use of rank correlation on WS353 mean raters scores as an instrument, smooths the noise from this imprecision and makes it appear to be precise.  The analysis shows that \code{95\%} of the pairs are statistically equivalent to \code{10\%} of their nearest neighbors, making the rank nearly meaningless. \textit{Any improvement in rank correlation displayed by one system over another is statistically accidental.}

\subsection{Reproducibility (of results of measurements)} 
\textit{The total measurement precision, including the components of variability that occur in the long term, and occurring from one measurement instrument to another. This is the closeness of the agreement between the results of measurements of the same measurand carried out under changed conditions of measurement. It requires specification of the conditions changed, e.g. principle of measurement, method of measurement, observer, measuring instrument, reference standard, location, conditions of use or time.} 

Crowd-powered instruments are rarely reproduced. However, when they are reproduced this typically involves collecting the crowd data again and compare with the original instrument. Thus, we re-ran the WS353 set of word pairs with Amazon MTurk using the original guidelines and, as before, collecting 13 judgements per pair for the full 353 set. Unlike the original, we collected work from 50 different raters to yield the same total number of judgements.

We measure the reproducibility of WS353 in two ways: IRR and rank correlations between the three datasets, shown in Tab.~\ref{tab:reproducibility_353}.

\begin{table}
\centering
\begin{tabular}{r|c|c|c}
    \hline
    \textbf{Sys} & \textbf{WS353} & \textbf{WS353r} & \textbf{WS353r2} \\
    \hline
    WS353 & 0.739 & 0.643 & 0.563 \\
    WS353r & 0.784 & 0.706 & 0.635 \\
    WS353r2 & 0.810 & 0.715 & 0.642 \\
    IRR ($\alpha$) & 0.59 & 0.59 & 0.52 \\
    \hline
\end{tabular}
\caption{Spearman's rank correlation between the original WS353 worker means and two subsequent reproductions of the WS353 measurements, as well as the Krippendorf-$\alpha$ for each. The $\alpha$ values are consisent, but the correlations are suprisingly low.}
\label{tab:reproducibility_353}
\end{table}

The rater means for each pair across the two data sets, WS353 and WS353r follow an exponential distribution, and share a Spearman $\rho$ of \code{0.87}.  The standard deviations are entirely different, \textit{resulting in an instrument with a lower precision}.  The mean standard deviation of WS353r is \code{2.06} (compared to WS353 at \code{1.7}), and a standard deviation of deviation of \code{.69}.  The standard deviations between the two sets are uniformly distributed and do not correlate well, with a Pearson of \code{0.26}.  The largest change in deviation between the two sets was the pair \code{(Maradona - football)}, which had a new standard deviation of \code{3.56} compared to \code{1.13} in the original.  This is likely due to the passage of time since the original was gathered.  Many of the pairs with the biggest changes are named entities, some of whom have changed or retired from popularity. The pair with the largest change in mean similarity is \code{(Arafat - peace)}, which had a new mean of \code{1.19} compared to \code{6.73} in the original.  The pair with the smallest change is \code{(king - cabbage)}, which stayed at a low \code{0.23}, although \code{(situation - conclusion)} stayed close to its original \code{4.81} with a \code{4.80}. The only named entities in the 50 least changed pairs are \code{(Harvard - Yale)}.  Surprisingly, several high variance tuples saw no change in their variance, \code{(precedent - information)} kept its stdev of \code{2.57}, as did \code{(morality - importance)} at \code{2.48}, though they both changed slightly in mean. 

Like precision, reproducibility appears to be a per-item measure for this instrument, but aggregate measures, as reported above, are telling.  The distribution and shape parameters can help us understand how much changes. Since standard deviation features prominently in the precision of the instrument, the reproducibility measure should include both, as with the analysis above.  

We don't have a strong baseline of measurement to understand how good or bad the reproducibility scores for the instrument are. We can imagine variations on reproducibility tests that may generate larger changes, such as using raters from a specific nation, or requiring some specific background or expertise. 

\subsection{Resolution of a measuring system} 
\textit{Resolution is the smallest change in a quantity being measured that causes a perceptible change in the corresponding indication, the smallest change the instrument can detect. In instruments with digital displays, it is the number of significant digits of a measurement system that can be meaningfully interpreted.} 
In crowd-powered instruments we can measure sensitivity through the variance per item, on the one hand, and by creating a new instrument (a variant of the original one). 

As with other measures, the absence of a true value makes resolution difficult to specify, but since it is about change, we can devise a second instrument, call it WS353res, capable of measuring the resolution of WS353.  WS353res is a crowd-powered instrument (which we will, in the interest of space, only briefly characterize) in which raters are presented with a pair of WS353 word pairs (a pair of pairs), and asked in which pair the words are more similar to each other.  

The principle of measuring is that, e.g. the pair \code{(computer - keyboard)} and the pair \code{(planet - sun)} are indistinguishably similar, even though they have different mean scores in WS353, while for \code{(baseball - season)} and \code{(media - gain)}, the former is more similar than the latter.  We expect that people will be able to distinguish some pairs from each other, and not be able to distinguish others.  The procedure was to sample 200 pairs of pairs from WS353 such that the sample was uniformly distributed by their difference in WS353 mean scores.  We again used 13 raters per pair of pairs, to stay aligned with WS353, and asked them to indicate the more similar pair or indicate that they were equally similar, e.g. for the pairs \code{(theater - history)} and \code{(announcement - news)} raters had to choose one of the following statements:
\begin{itemize}
    \item \code{Theater:history} are more similar to each other than \code{announcement:news}
    \item \code{Announcement:news} are more similar to each other than \code{Theater:history}
    \item The pairs are equally similar
\end{itemize}
We then calculated the agreement-weighted decision for each pair of pairs.  For example, given \code{(computer - keyboard)} and \code{(planet - sun)}, \code{80\%} of the raters voted these pairs to be equally similar, whereas for \code{(baseball - season)} and \code{(media - gain)}, \code{92\%} of the raters voted the former more similar.

\begin{figure}[hbt!]
\centering
\includegraphics[width=0.9\columnwidth,height=150pt]{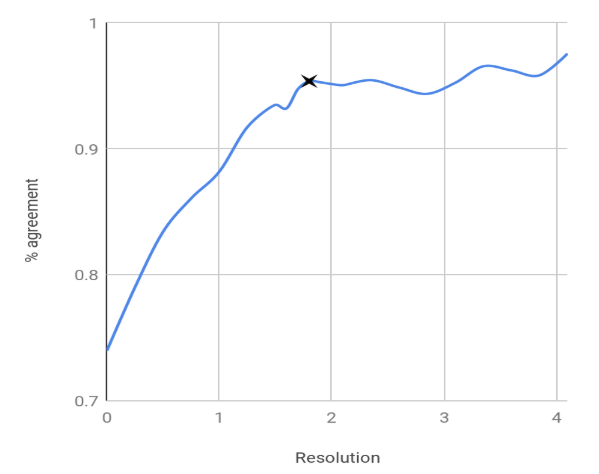}
\caption{Agreement between the WS resolution (which corresponds to the difference in WS353 scores) instrument and WS353 at different thresholds of distance between pairs. At a resolution of 1.8, there is 95\% agreement.}
\label{fig:resolution}
\end{figure}

Finally, to use the WS353res instrument to measure WS353's resolution, we aligned these 200 pairs of pairs with their WS353 distances, and measured the agreement between the two instruments at different distance thresholds.  For example, \code{(computer - keyboard)} and \code{(planet - sun)} differ by \code{0.4} in WS353, and would count as disagreement at any threshold $\leq 0.4$, and would be ignored at thresholds above that.  For \textit{(baseball - season)} and \code{(media - gain)}, they differ in WS353 by \code{3.09} in favor of the former, and would be considered agreement for thresholds $\leq 3.09$.  As shown in Figure \ref{fig:resolution}, the threshold where we reach \code{95\%} agreement between the two instruments is \code{1.8}. In other words, at pairwise distances below \code{1.8}, there is a greater that \code{5\%} chance two pairs are indistinguishably similar, and thus interchangeable in rank. The probability of indistinguishability increases to \code{25\%} with no threshold.  We call the sensitivity of the instrument \code{1.8} at \code{95\%}.

It is not immediately clear how to use the instrument resolution. What it tells us is that pairs of word pairs that differ by less than the resolution score may be equivalent in similarity - the instrument cannot tell them apart. There are \code{61,425} (unique) pairwise distances in WS353, and less than half of them, \code{33,359}, are above the sensitivity threshold. Similar to the precision, it indicates that the rank order can be quite arbitrary, as large numbers of pairs are within \code{1.8} of each other and may be interchangeable in rank. To address this, we formed a matrix whose rows and columns are the word pairs and whose cells are the distances between them.  The upper triangle of this matrix consists of those \code{61,425} values, representing the pairwise distances between all word pairs. These values are normally distributed and can be used, after normalization, as the basis of a correlation with the same pairwise distance from a system to be tested.

To account for the instrument resolution, we identify all the cells whose values are below the resolution and remove them from the set. It is expected, and analysis confirms it, that the above resolution distances are an easier target to correlate with.  

Despite its lower precision, the WS353r instrument has very similar resolution to WS353 at \code{1.85}.



\section{Implications}
Many studies rely on WS353\footnote{\url{https://aclweb.org/aclwiki/WordSimilarity-353_Test_Collection_(State_of_the_art)}} to evaluate their performance, e.g. \cite{Halawi2012,luong2013} to mention some. There are other similar datasets, such as \cite{Rubenstein1965} and \cite{Li2006}, as used by \cite{Hassan2011}. Most of these studies compute rank correlation between the mean human judgment as reported in the dataset and the system's score. That is, they are taking the WS353 instrument as a perfect instrument, and ignoring aspects of precision, repeatability, reproducibility, sensitivity, and resolution of the instrument itself. \textit{Taking the instrument characteristics into account might alter the interpretation of results of these systems.} 

We obtained data, in particular the system predictions on each pair, from published studies with WS353 by \cite{Halawi2012}, which we utilize in the subsequent analyses. \cite{Halawi2012} proposes a large-scale data mining approach to learning word-word relatedness, called CLEAR. The authors measured the performance of CLEAR and CLEAR+TSA using \textit{WS353+}$\rho$ as the instrument, and compared to previously published approaches ESA and TSA \cite{Radinsky2011} measured with the same instrument.  The comparison is intended to show a significant improvement with the new methods.

\subsection{Taking Precision into Account}
The published WS353 data contains thirteen repetitions per word-pair, obtained by presenting the same pair to different raters. The \textit{WS353+}$\rho$ instrument is based on the (Spearman's $\rho$) rank correlation between system predictions and the rater means for each pair.  As we have discussed, there is variability across repetitions within each word pair, quantified as the instrument's precision per pair.  For example, the word pair \code{(precedent - antecedent)} has a mean similarity score of \code{6.04}, but a standard deviation of \code{3.01}; some raters report it is extremely similar (score = \code{9.5}, compared to \code{10} for \code{(tiger - tiger)}) and some raters report no similarity (score = \code{0}).  This variation should affect the confidence in the evaluations made using it. 

In Table \ref{tab:implications_figure1}, we show summary statistics for each system's performance by computing the rank correlation $\rho$ \textit{per rater}.  The first observation is that there is a huge variance across repetitions, the delta $\rho$ between the least and most correlated raters being around \code{0.4} and the standard deviation across repetitions larger than \code{0.1}.  Second, each system's per-rater-$\rho$ distributions lie well within 1 standard deviation of each other. Finally, while the mean per-rater-$\rho$ scores preserve the same rankings as the reported WS353-$rho$, the error introduced by the item precision makes it clear that \textit{the instrument is not capable of measuring the difference between these four systems}.

\begin{table}[]
    \centering
    \begin{tabular}{r|r|r|r|r|r}
         \hline
         \textbf{exp} & \textbf{min} & \textbf{max} & \textbf{mean} & \textbf{stdev} & \small \textbf{WS353-$\rho$}  \\
         \hline
         \small ESA  & \small 0.263 & \small 0.632 & \small 0.498 & \small 0.108 & \small 0.739 \\
         \small TSA  & \small 0.268 & \small 0.644 & \small 0.519 & \small 0.114 & \small 0.784 \\
         \small CLEAR  & \small 0.294 & \small 0.704 & \small 0.544 & \small 0.127 & \small 0.810 \\
         \small CLEAR+TSA & \small 0.293 & \small 0.703 & \small 0.562 & \small 0.125 & \small 0.841 \\
         \hline
    \end{tabular}
    \caption{Effect of WS353 precision on measuring four systems. The min, max, mean, stdev of each system's $\rho$ score vs. each WS353 rater, and the originally reported $\rho$ vs. the rater means. The four systems' scores are not significantly different.}
    \label{tab:implications_figure1}
\end{table}

To validate this conclusion, we performed a two-tailed t-test for the null hypothesis for every pair of systems that they have identical expected values across the raters (Table \ref{tab:system_t_tests}). Given that the p-values lie between \code{0.19} and \code{0.72}, we fail to reject the null hypothesis for each of the six pairs of systems. That is, \textit{the} WS353+$\rho$ \textit{instrument lacks the precision to measure the difference between these four systems}, and we can't say for sure that they are any different. 

\begin{table}[]
\centering
\begin{tabular}{r|l|r|r}
\hline
\textbf{System 1} & \textbf{System 2} & \textbf{t-statistic} & \textbf{p-value}\\
\hline
\small CLEAR+TSA & \small ESA  & \small 1.34 & \small 0.19 \\
\small CLEAR+TSA & \small TSA  & \small 0.87 & \small 0.39 \\
\small CLEAR+TSA & \small CLEAR & \small 0.36 & \small 0.72 \\
\small ESA  & \small TSA & \small -0.47 & \small 0.64 \\
\small ESA  & \small CLEAR & \small -0.95 &	\small 0.35 \\
\small TSA  & \small CLEAR & \small -0.49 & \small 0.63\\
\hline
\end{tabular}
\caption{Pairwise significance tests between the per-rater $\rho$ scores.}
\label{tab:system_t_tests}
\end{table}


\subsection{Taking Resolution into Account}
Table \ref{tab:resolution_table} shows the Pearson correlation between: the $l2$-normalized distances between all WS353 pairs and the $l2$-normalized distances between all the experiment pairs. Each column of the table represents the correlations at a particular instrument resolution, such that the correlated distances only include those on pairs for which the ws353 distance is above the resolution. The counts show how many pairs are included at that level of resolution.

We were expecting the results to show that, as the resolution threshold increases, the four systems become decreasingly different.  The results do not show this, however they do lend more evidence to support the hypothesis that measuring the differences between these four systems is beyond the \textit{WS353+}$\rho$ instrument's capabilities.  In particular, the results show that CLEAR, and not CLEAR+TSA, has a higher correlation with the WS353 pairwise distances.

This experiment has high statistical power because of the number of pairs (bottom row in Table \ref{tab:resolution_table}), and these differences are significant for that reason. 

\begin{table}[]
\centering
\begin{tabular}{r|r|r|r|r|r}
\hline
 & \textbf{0.0} & \textbf{0.9} & \textbf{1.8} & \textbf{2.7} & \textbf{3.6}\\
\hline
\small CLEAR & \small 0.742  & \small 0.782 & \small 0.816 & \small 0.845 & \small 0.867 \\
\small CLEAR+TSA & \small 0.651  & \small 0.675 & \small 0.697 & \small 0.723 & \small 0.744 \\
\small TSA  & \small 0.576  & \small 0.599 & \small 0.621 & \small 0.646 & \small 0.668 \\
\small ESA  & \small 0.506  & \small 0.530 & \small 0.553 & \small 0.577 & \small 0.598 \\
\hline
\small Count  & \small 61425 & \small 46375 & \small 33359 & \small 23280 & \small 15693\\
\hline
\end{tabular}
\caption{Pearson correlation between (1) the normalized distances between all ws353 pairs and (2) the normalized distances between all the predicted pairs, at different values of instrument resolution.}
\label{tab:resolution_table}
\end{table}


\subsection{Taking Reproducability into Account}
Finally, we measured the four systems against the two reproduced instruments, WS353r and WS353r2, the results are shown in Tab.~\ref{tab:reproducibility_exp}.  The relative ordering of the systems under rank corrlation remains unchanged across the three instruments, although the actual $\rho$ values change dramatically.  At best we can conclude from this that reproducability confirms the measurements, at worst the raw values show the lowest performing system according to WS353 performs better than the highest performing system according to WS353r2, lending some doubt to the ranking.  More precise conclusions are left to future work.

\begin{table}
\centering
\begin{tabular}{r|c|c|c}
    \hline
    \textbf{Sys} & \textbf{WS353} & \textbf{WS353r} & \textbf{WS353r2} \\
    \hline
    ESA & 0.739 & 0.643 & 0.563 \\
    TSA & 0.784 & 0.706 & 0.635 \\
    CLEAR & 0.810 & 0.715 & 0.642 \\
    CLEAR+TSA & 0.841 & 0.748 & 0.670 \\
    \hline
\end{tabular}
\caption{Spearman's rank correlation of each system's predictions to the mean rater scores of the original and two subsequent reproductions of the WS353 measurements.  While the relative order of the system scores remain the same within each reproduced measurement, the scores change by far more than the differences between systems.}
\label{tab:reproducibility_exp}
\end{table}

\section{Discussion}

This paper is the first step towards standardized characterization of crowd-powered instruments, such as benchmark datasets. The main contribution is in adapting metrology as a framework for characterizing the quality and reliability of crowd-powered instruments. Crowd-powered benchmarks have a great impact on the state-of-the-art both AI and Human Computation, as they serve as a compass to guide progress in the field, and as such we need to be able to understand their characteristics as an instrument: 
\begin{itemize}
    \item \textbf{AI research: } Crowd-powered instruments have been highly effective in being the north star for research by providing a shared measure of progress. Ignoring the instrument characteristics is perilous, and could result in wasted research cycles chasing improvements that might be well within the precision, resolution, and sensitivity of the instrument. 
    \item \textbf{Human Computation research:} Characterizing the crowd-based methodologies will tell us how we can build better instruments and justify allocating resources to improving them. The job of task design for data collection is often relegated to ad hoc templates and guidelines, which reduces the reusability of the data collected. We believe that this metrological approach offers an empirical framework for characterizing and improving crowd-powered instruments. 
\end{itemize}

In addition to characterizing the widely used instruments, there are many open questions around how to better characterize crowd powered instruments.  We also need to look toward the empirical and theoretical body of work from behavioural sciences around the principles of measurements involving humans \cite{Berglund2013}. All the data from the experiments in this paper will be released for further experimentation. 




\bibliography{Crowd_instrument}
\bibliographystyle{aaai}

\end{document}